\documentclass[sigconf]{acmart}

\usepackage{bm}

\AtBeginDocument{%
  }

\acmSubmissionID{498}

\usepackage{algorithm}
\usepackage{booktabs}
\usepackage[normalem]{ulem}
\useunder{\uline}{\ul}{}
\usepackage{multirow}
\usepackage{amsmath}
\usepackage{color}
\usepackage{algpseudocode}  

\usepackage{flushend}
\usepackage{balance}

\begin{document}

\title{Human-Centric Topic Modeling with Goal-Prompted Contrastive Learning and Optimal Transport}


\author{Rui Wang$^{\dagger}$, Yi Zheng$^\dagger$, Dongxin Wang$^{\dagger}$, Haiping Huang$^{\dagger*}$, Yuanzhi Yao$^{\ddagger}$, Yuxiang Zhou$^\diamondsuit$, Jialin Yu$^{\heartsuit}$, Philip Torr$^{\heartsuit}$}
 
\affiliation{%
 \institution{$^\dagger$School of Computer Science, Nanjing University of Posts and Telecommunications   \\
 $^\ddagger$School of Computer Science and Information Engineering, Hefei University of Technology\\
 $^\diamondsuit$School of Electronic Engineering and Computer Science, Queen Mary University of London \\
 $\heartsuit$ Department of Engineering Science, University of Oxford\\
 email:\{rui\_wang, 1223045708, 1224045815, hhp\}@njupt.edu.cn, yaoyz@hfut.edu.cn, yuxiang.zhou@qmul.ac.uk, \{jialin.yu, philip.torr\}@eng.ox.ac.uk
 }
 \country{}
}

 \thanks{* Corresponding author.}

\renewcommand{\shortauthors}{Anonymous Author, et al.}
\renewcommand{\shortauthors}{Trovato et al.}

\setlength{\abovedisplayskip}{1.3pt} 
\setlength{\belowdisplayskip}{1.3pt}
\vspace{-0.2em}
\vspace{-0.2em}

\renewcommand\floatpagefraction{.9}
\renewcommand\dblfloatpagefraction{.9} 
\renewcommand\topfraction{.9}
\renewcommand\dbltopfraction{.9} 
\renewcommand\bottomfraction{.9}
\renewcommand\textfraction{.1}   
\setcounter{totalnumber}{50}
\setcounter{topnumber}{50}
\setcounter{bottomnumber}{50}



\begin{abstract}

Existing topic modeling methods, from LDA to recent neural and LLM-based approaches, which focus mainly on statistical coherence, often produce redundant or off-target topics that miss the user’s underlying intent. We introduce ``Human-centric Topic Modeling'' (\emph{Human-TM}), a novel task formulation that integrates a human-provided goal directly into the topic modeling process to produce interpretable, diverse and goal-oriented topics. To tackle this challenge, we propose the \textbf{G}oal-prompted \textbf{C}ontrastive \textbf{T}opic \textbf{M}odel with \textbf{O}ptimal \textbf{T}ransport (GCTM-OT), which first uses LLM-based prompting to extract goal candidates from documents, then incorporates these into semantic-aware contrastive learning via optimal transport for topic discovery. Experimental results on three public subreddit datasets show that GCTM-OT outperforms state-of-the-art baselines in topic coherence and diversity while significantly improving alignment with human-provided goals, paving the way for more human-centric topic discovery systems.

\end{abstract}



\maketitle

\section{Introduction}

\begin{figure}[!t]
    \centering
    \includegraphics[width=0.9\linewidth]{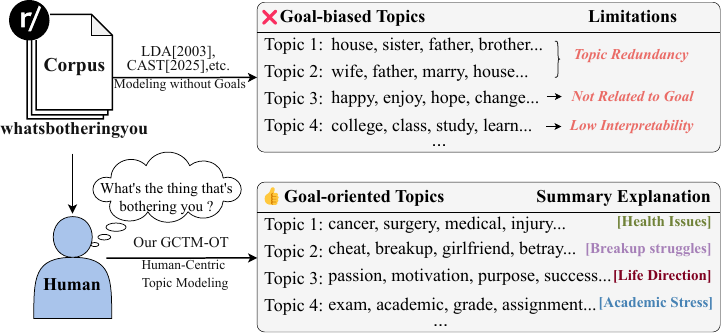}
    \caption{Key differences between Traditional Topic Modeling and Human-Centric Topic Modeling (with human's goal).}
    \vspace{-0.2cm}
    \label{fig:XXX}
\end{figure}

Topic models~\cite{DBLP:journals/is/AbdelrazekEGMH23,DBLP:journals/air/WuNL24}, such as the Latent Dirichlet Allocation (LDA)\cite{DBLP:journals/jmlr/BleiNJ03} and advanced variants including the Corpus-Aware Self-similarity enhanced Topic Model (CAST)~\cite{DBLP:conf/naacl/MaXYVHLN25}, have been widely adopted for open information extraction~\cite{DBLP:conf/emnlp/WangZH19,DBLP:journals/tnn/WangZHZ25} and knowledge discovery~\cite{DBLP:conf/kdd/WanSJKCNSSWYABJ24,DBLP:journals/eswa/YuX23}. These approaches uncover coherent semantic patterns from large text corpora without supervision. However, when the analytical objective is guided by \textbf{a specific human goal}, these methods often fall short because they are not designed to adapt topic discovery to user-defined intentions without extensive manual intervention. Consider the example in Figure~\ref{fig:XXX}, where the human goal, expressed in question \emph{"What’s the thing that's bothering you?"}, is to identify the main issues of concern within a collection of text responses. Traditional topic models, which are optimized solely for statistical coherence, often produce topics that, although internally consistent, are redundant, less interpretable, or unrelated to the goal. For example, topics containing terms such as \emph{`house'}, \emph{`sister'}, \emph{`father'} or \emph{`happy'}, \emph{‘enjoy’}, \emph{`hope'} are coherent in a statistical sense but do not address the intended analytical objective. As a result, important patterns such as \emph{`breakup struggles'}, \emph{`health issue'} or \emph{`academic stress'} may be overlooked or obscured, requiring substantial manual filtering and interpretation.

To address these limitations, we propose a new task formulation called Human-centric Topic Modeling (\emph{Human-TM}). As illustrated in Figure~\ref{fig:XXX}, given a text corpus (e.g. Subreddit corpus \emph{`Whatsbotheringyou'})~\footnote{\url{https://www.reddit.com/r/whatsbotheringyou/}} and a human-provided free-text goal (e.g. \emph{``What’s the thing that's bothering you?''}), \emph{Human-TM} aims to automatically generate a set of goal-oriented topics. These topics should satisfy the following three requirements: 1). Topics are semantically relevant to the human-provided goal. 2). Topics are semantically coherent and easily interpretable. 3). Topics are diverse (non-redundant) and collectively provide broad coverage of the corpus.

To tackle the \emph{Human-TM} task, we propose the \textbf{G}oal-prompted \textbf{C}ontrastive \textbf{T}opic \textbf{M}odel with \textbf{O}ptimal \textbf{T}ransport (GCTM-OT), which formulates topic modeling as a representation learning procedure based on advanced prompt engineering~\cite{DBLP:journals/csur/EdemacuW25} and optimal transport theory~\cite{DBLP:journals/ftml/PeyreC19}. The core idea is to incorporate goal information, identified through LLM-based prompting from texts, into the modeling process for goal-oriented topic extraction. Specifically, GCTM-OT first generates phrase-style goal candidates for each document using a goal-oriented summarization prompt and aggregates them into a corpus-level goal set. To improve topic interpretability, it integrates contextualized word representations from a transformer~\cite{DBLP:conf/naacl/DevlinCLT19} to build contextualized document representations and models topics with topic representations to infer document-topic distributions. Moreover, to embed goal information and encourage topic diversity, it forms goal representations by clustering goal candidates in the corpus-level goal set and employs optimal transport~\cite{DBLP:journals/ftml/PeyreC19} to guide topic representation training. After training with the supervision signals provided by contrastive learning and optimal transport, GCTM-OT is capable of identifying interpretable and diverse goal-oriented topics. Additionally, it could generate a phrase-style topic summary for each topic, offering document-level topic interpretation.

The main contributions of this work are summarized as:
\vspace{-0.1cm}

\begin{itemize}
    \item We introduce \emph{Human-TM}, a novel topic modeling formulation that takes human goals into account and generates interpretable, diverse and goal-oriented topics.
    \item We propose the novel \textbf{G}oal-prompted \textbf{C}ontrastive \textbf{T}opic \textbf{M}odel with \textbf{O}ptimal 
    \textbf{T}ransport (GCTM-OT), which leverages LLM-based prompting to identify goal candidates within texts and incorporates them into the modeling process via contrastive learning and optimal transport, enabling goal-oriented topic extraction.
    \item We evaluate GCTM-OT on three Subreddit datasets. Experimental results show that GCTM-OT outperforms state-of-the-art baselines in topic coherence and diversity. In addition, it generates topics that closely align with human goals, as measured by {\color{black}Goal Similarity ($GS$), Goal-relevant Topic Rate ($GTR$)} and Goal Coverage Rate ($GCR$) metrics.
\end{itemize}

\section{Related Work}
Our work is relevant to neural topic modeling and Large Language Model based prompting.

\subsection{Neural Topic Modeling}
Modeling topics using generative neural networks, such as Variational Autoencoders (VAE)~\cite{DBLP:journals/corr/KingmaW13} and Generative Adversarial Networks (GAN)~\cite{DBLP:conf/nips/GoodfellowPMXWOCB14}, has emerged as an active area of research.

Miao et al. proposed the Neural Variational Document Model (NVDM)~\cite{DBLP:conf/icml/MiaoYB16}, which is the pioneering work in this field, based on the VAE. Likewise, Wang et al. proposed the Adversarial-neural Topic Model (ATM)~\cite{DBLP:journals/ipm/WangZH19} and the Bidirectional Adversarial Topic (BAT)~\cite{DBLP:conf/acl/WangHZHXYX20} model in an adversarial manner and utilized a generator network to capture semantic patterns. Inspired by NVDM, Xu et al. proposed the VONT~\cite{DBLP:conf/acl/XuJRIZ23} by employing a mixture of von-Mises Fisher distributions as topic prior. Ma et al. utilized the word embeddings to filter out irrelevant words and proposed the Corpus-Aware Self-similarity Topic (CAST)~\cite{DBLP:conf/naacl/MaXYVHLN25} model. Fang et al. incorporated the dynamic word representations into the modeling process and proposed the Contextualized Word Topic Model (CWTM)~\cite{DBLP:conf/coling/00110P24}. To mine the public's opinion towards ChatGPT, Wang et al. employed a topic disentangle mechanism and proposed the Disentangled Contextualized Topic Model (DisCTM)~\cite{DBLP:conf/wsdm/WangLWCYH25}. However, all of these approaches are not able to incorporate the human's goal into the modeling process, resulting in generic topics that do not match specific human needs.

\subsection{Large Language Model based Prompting}

 Advances in Large Language Models (LLMs) have made them a thriving research topic and have been explored for their robust capabilities in NLP tasks~\citep{DBLP:conf/emnlp/0001WS23,DBLP:conf/emnlp/StrongA024,DBLP:conf/emnlp/WangSZ23}. 

Meanwhile, Prompt Engineering~\citep{DBLP:journals/corr/abs-2402-07927} has emerged as an indispensable approach to extend the capability of LLMs. Firstly, Radford et al. proposed the Zero-shot prompting~\citep{radford2019language} to offer a paradigm shift in leveraging LLMs. By providing several input-output examples to induce the target of the given task, Brown et al. proposed the Few-shot Prompting~\citep{brown2020language}. Along this line, Wei et al. introduced the Chain-of-Thought (CoT)~\citep{DBLP:conf/nips/Wei0SBIXCLZ22} prompting to guide LLMs and form a step-by-step reasoning scheme. Recently, scholars have also explored using prompting for topic extraction. Pham et al. employed the LLM-prompting and proposed the TopicGPT~\cite{DBLP:conf/naacl/PhamHSRI24} for topic phrase generation. Mu et al. proposed the Large Language Model Topic Extraction (LLM-TE)~\cite{DBLP:conf/coling/MuDBS24}, which is an automatic topic generation framework based on LLM. Chang et al. proposed to improve topic quality by designing a topic refinement mechanism~\cite{chang2024large} with LLM-based prompting. 

\section{\emph{Human-TM} Task Formulation}

Given a collection of $N$ documents $D\!=\!\{x_1, x_2,...,x_N\}$ and a human-provided goal description $\mathcal{H}_g$, \emph{human-TM} aims to mine a set of $K$ topics $T\!=\!\{t_1, t_2,...,t_K\}$ that satisfy the following requirements:
\begin{enumerate}
    \item Goal-Oriented: Each topic $t_k$ ($k\!\in\!\{1,2,...,K\}$) must be semanticlly related to human's goal. For example, if the goal $\mathcal{H}_g$=\emph{``What's the thing that's bothering you?''}, extracted topics should be \emph{``Breakup Struggles''}, \emph{``Acdamic Stress''} and etc. 
    \item Interpretable: Each topic $t_k$ ($k\in\{1,2,...,K\}$) must be represented by a list of semantically coherent words $t_k^w$ and a human-readable summary explanation $t_k^s$, such as [\emph{`cheat'}, \emph{`breakup'}, \emph{`girlfriend'}, \emph{`betray'}, \emph{`ring'}, \emph{`break'}, \emph{`divorce'}, \emph{`relationship'}, \emph{`partner'}, \emph{`devastate'}] and \emph{``breakup struggles''}. 
    \item Diverse and Maximum Coverage: The topics in $T$ should be semantically diverse, and their union should cover the majority of goal candidates retained in the corpus. 
\end{enumerate}

\section{Methodology}

\begin{figure*}[!htbp]
    \centering
    \includegraphics[width=0.94\linewidth]{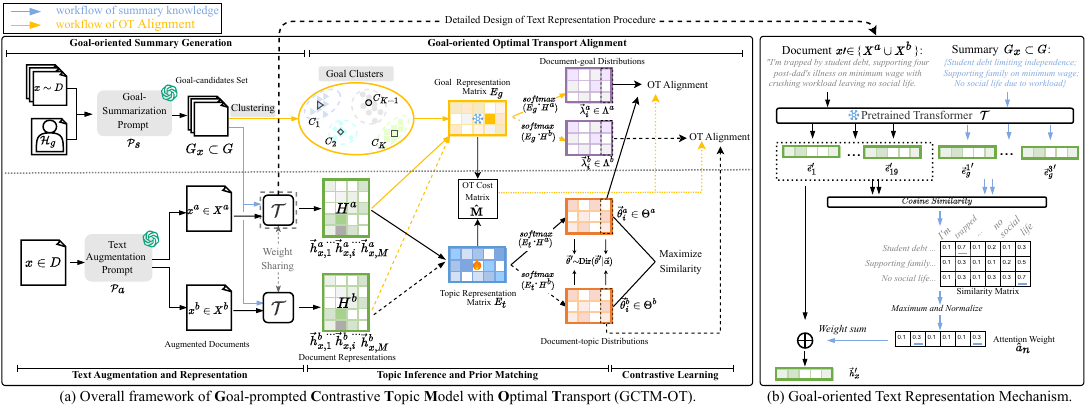}
\vspace{-0.4cm}
    \caption{The framework of GCTM-OT (a) and details of Goal-oriented Text Representation Mechanism (b).}
    \label{fig:model}
\end{figure*}

As illustrated in Figure~\ref{fig:model} (a), our proposed Goal-prompted Contrastive Topic Model with Optimal Transport (GCTM-OT) comprises five components: 1). \emph{Goal-oriented Summary Generation (top-left):} Extracts human-goal–related phrases from each document so that later topic discovery stays focused on the target intent.  It uses the goal summarization prompt $\mathcal{P}_{s}$ to generate $G_x$ for each document $x$, which are then aggregated into the corpus-level goal set $G$. 2). \emph{Text Augmentation and Representation (bottom-left):} Produces diverse yet semantically consistent views of each document so that contrastive learning is effective. Using the text augmentation prompt $\mathcal{P}_a$, it generates pair $(x^a, x^b)$, then integrates goal information from $G_x$ into a transformer $\mathcal{T}$ to produce goal-oriented representations $(\vec h^a_x, \vec h^b_x)$. 3). \emph{Topic Inference and Prior Matching (bottom-middle):} Maps goal-aware document representations to topic distributions while maintaining interpretability. It models topics using the trainable topic representation matrix $E_t$ and infers the document-topic distributions $\vec \theta^a$ and $\vec \theta^b$ for $x^a$ and $x^b$, respectively. These distributions are then matched to a Dirichlet prior, leveraging its multiple-peak property to improve topic interpretability. 4). \emph{Semantic-aware Contrastive Learning (bottom-right):} Encourages topic representations to capture distinctive semantic boundaries between documents. It optimizes $E_t$ through contrastive learning, bringing semantically similar document pairs closer and separating unrelated pairs in the learned topic space. 5). \emph{Goal-oriented Optimal Transport Alignment (top-right):} Aligns the learned topics with human goals to ensure relevance and diversity of topics. It first clusters the corpus-level goal set $G$ to form a goal representation matrix $E_g$, infers document–goal distributions $(\vec \lambda^a, \vec \lambda^b)$, and then applies optimal transport to align the document–topic distributions $(\vec \theta^a, \vec \theta^b)$ with the corresponding document–goal distributions $(\vec \lambda^a, \vec \lambda^b)$, directly injecting goal structure into topic learning. Meanwhile, Figure \ref{fig:model} (b) illustrates the details of the goal-oriented text representation mechanism. The functionalities of these components are described in detail below.

\subsection{Goal-oriented Summary Generation}

\begin{figure}[!t]
    \centering
    \includegraphics[width=0.85\linewidth]{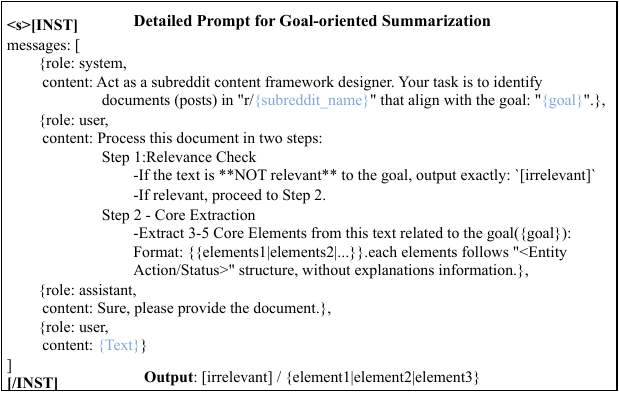}
    \caption{Details of Goal Summarization Prompt $\mathcal{P}_s$.}
    \label{fig:prompt1}
    \vspace{-0.3cm}
\end{figure}

To ensure semantic alignment between the mined topics and the human-provided goal $\mathcal{H}_g$, we first identify and extract potential goal candidates from the corpus. For each document $x$ in the corpus $D$, we use an LLM~\cite{fui2023generative} with a goal-summarization prompt $\mathcal{P}_s$ to generate document-specific goal candidates $G_x=\{g_x^1, g_x^2,...,g_x^{N_{g_x}}\}$. Here, $N_{g_x}$ denotes the number of goals in document $x$, ranging from 3 to 5 in our experiments. If a document is deemed irrelevant to the goal $\mathcal{H}_g$, it is annotated as \emph{`irrelevant'} and excluded from the corpus. The details of the prompt $\mathcal{P}_s$ are shown in Figure~\ref{fig:prompt1}, which follows the OpenAI-released API guidelines\footnote{\url{https://platform.openai.com/docs/guides/prompt-engineering}\label{fn:opanai}}.

We then aggregate document-specific goal sets $G_x^n$ ($n\! \in\! \{\!1\!,\!2,\!\dots,\!N\}$) to construct the corpus-level goal set $G = G_x^1 \cup G_x^2 \cup \dots \cup G_x^N$, which is later used to produce the goal representation matrix $E_g$ (Section~\ref{sec:goalrepresentation}).

\begin{figure}[!t]
    \centering
    \includegraphics[width=0.85\linewidth]{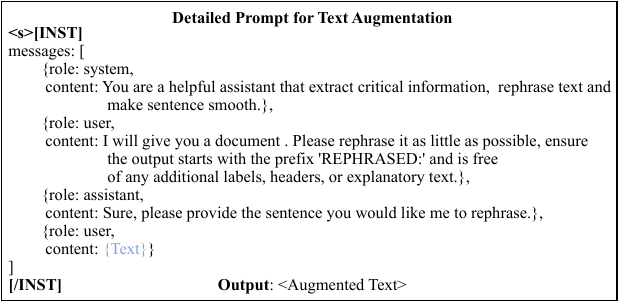}
    \vspace{-0.3cm}
    \caption{Details of Text Augmentation Prompt $\mathcal{P}_a$.}
    \label{fig:prompt2}
       \vspace{-0.2cm}
\end{figure}

\subsection{Text Augmentation and Representation}

To ensure that the augmented documents preserve the semantic meaning of the original documents, which is critical for effective contrastive representation learning~\cite{DBLP:conf/icml/0001I20}, we propose using a text augmentation prompt $\mathcal{P}_a$ to generate the augmented documents.

Specifically, for each document $x$ in the batch $X$, we employ a text augmentation prompt $\mathcal{P}_a$ to generate semantically consistent documents $x^a$ and $x^b$ by rephrasing $x$ using a large language model. The details of the augmentation prompt $\mathcal{P}_a$ are illustrated in Figure~\ref{fig:prompt2}, which also follows the OpenAI released API guideline\textsuperscript{~\ref{fn:opanai}}.

Then, for each augmented document $x'\! \in\! \{x^a, x^b\}$, we generate a goal-oriented contextualized representation $\vec h'_{x}$ using a transformer~\cite{DBLP:conf/naacl/DevlinCLT19} $\mathcal{T}$ (Figure~\ref{fig:model}b). This model integrates the contextual information from the document with its corresponding goal candidates $G_x\!=\!\{g_x^1,g_x^2,\dots,g_x^{N_g}\}$. Specifically, we first represent the document $x' \!=\! [w'_1, w'_2, \dots, w'_{N_x'}]$ using transformer $\mathcal{T}$:
\begin{equation}
    [\vec e'_1,\vec e'_2,...,\vec e'_{N_x'}]=\mathcal{T}([w'_1,w'_2,...,w'_{N_x'}])
    \label{c1}
\end{equation}
where $\vec e'_n$ ($n\in\{1,2,...,N_{x}'\}$) denotes the contextualized word representation of the $n$-th word in the document. Next, we encode the document-specific goal candidates with:
\begin{equation}
    [\vec e_{g}^{'1};\vec e_{g}^{'2};...;\vec e_{g}^{'N_{g_x}}]=[\bar{\mathcal{T}}(g_x^1); \bar{\mathcal{T}}(g_x^2);...;\bar{\mathcal{T}}(g_x^{N_{g_x}})]
    \label{c2}
\end{equation}
where $\vec e_{g}^{'i}$ ($i \in \{1, 2, \dots, N_{g_x}\}$) is the contextualized embedding of the $i$-th goal $g_x^{i}$, and $\bar{\mathcal{T}}(\cdot)$ denotes the mean of token embeddings produced by the model $\mathcal{T}$ for that goal phrase. Then, for the $n$-th ($n\in\{1,2,...,N_x'\}$) word $w'_n$, we could calculate its semantic correlations with each goal candidate and assign its attention weight $a_n$ as the maximum cosine similarity:
\begin{equation}
    a_{n}=\max([\cos(\vec e'_n,\vec e_{g}^{'1}),\cos(\vec e'_n,\vec e_{g}^{'2}),...,\cos(\vec e'_n,\vec e_{g}^{'N_g})])
    \label{c3}
\end{equation}
where $\cos(\cdot,\cdot)$ means the cosine similarity. The use of the maximum operation ensures that each word is associated with its most relevant goal, preserving the strongest semantic link without dilution from less-related goals. This highlights words with at least one strong goal alignment while ignoring weaker associations. Finally, we obtain the contextual representation $\vec h'_{x}$ of document $x'$ with:
\begin{align}
    [\hat{a}_1,\hat{a}_2,...,\hat{a}_{N_x'}]=&\textrm{softmax}([a_1,a_2,...,a_{N_x'}])\\
    \vec h'_{x}=&\sum\nolimits_{n=1}^{N_x'}\hat{a}_{n}\cdot \vec e'_{n}
    \label{c4}
\end{align}
where $\textrm{softmax}(\cdot)$ means the softmax normalization. Thus, we could generate contextualized document representations $\vec h^a$ and $\vec h^b$ for the LLM-augmented documents $x^a$ and $x^b$. By incorporating contextual and goal candidate information in this way, these representations emphasize goal-relevant semantics and facilitate the extraction of interpretable and goal-oriented topics.

\vspace{-0.2cm}
\subsection{Topic Inference and Prior Matching}
Given the contextualized document representations ($\vec h^a$ and $\vec h^b$) from the previous stage, this module aims to infer document-topic distributions ($\vec \theta^{a}$ and $\vec \theta^{b}$) for the augmented documents ($x^a$ and $x^b$). Additionally, it matches the inferred document-topic distributions to the Dirichlet prior~\cite{lin2016dirichlet}, leveraging the multiple peaks property of its density to improve topic interpretability~\cite{DBLP:conf/nips/WallachMM09}.

To capture semantic patterns in the contextualized document representations, GCTM-OT models topics using a trainable topic representation matrix $E_t\! \in\! \mathbb{R}^{K \times H}$, where $H$ denotes the dimensionality of the transformer representation. Specifically, for the augmented document $x^a$, GCTM-OT infers its document-topic distribution $\vec \theta^a$ with:
\begin{equation}
    \vec \theta^a=\textrm{softmax}([\cos(\vec h^a,\vec e_t^1),\cos(\vec h^a,\vec e_t^2),...,\cos(\vec h^a,\vec e_t^K)]) \label{eq:inference}
\end{equation}
where $\vec e_t^k$ ($k\!\in\!\{1,2,...,K\}$) represents the topic representation of the $k$-th topic. Likewise, the document-topic distribution of $x^b$ could be obtained by feeding it with $\vec h^b$.


Moreover, to ensure topic interpretability, we match the inferred document-topic distributions to the Dirichlet prior via Maximum Mean Discrepancy (MMD)~\cite{DBLP:journals/jmlr/GrettonBRSS12}. Specifically, given two batches of inferred document-topic distributions $\Theta\!=\!\{\vec \theta_{1}, \vec \theta_2,...,\vec \theta_{2M}\}\!=\!\Theta^a\cup \Theta^b\!=\!\{\vec \theta^a_1,\vec \theta^a_2,...,\vec \theta^a_M,\vec \theta^b_1,$ $\vec \theta^b_2,...,\vec \theta^b_M\}$ and two batches of random samples $\Theta'\!=\!\{\vec \theta'_1,\vec \theta'_2,...,\vec \theta'_{2M}\}$ drawn from the Dirichlet prior $\textrm{Dir}(\vec \theta'|\vec \alpha)$, we match the inferred document-topic distributions $\Theta$ to the $\Theta'$ with the formula:
\begin{equation}
    MMD(\Theta,\Theta')\!=\!\frac{1}{A}\sum_{i\neq j}[\textrm{k}(\vec \theta_i,\vec \theta_j)\!+\!\textrm{k}(\vec \theta'_i,\vec \theta'_j)]\!-\!\frac{1}{2M^2}\sum_{i,j}\textrm{k}(\vec \theta_{i},\vec \theta'_{j})
    \label{mmd}
\end{equation}
where $A\!=\!2M(2M-1)$, $M$ denotes the batch size, $\mathrm{k}(\cdot,\cdot)$ is the kernel function~\cite{terrell1992variable}, and $\vec \alpha$ is the hyperparameter of the Dirichlet prior.




\vspace{-0.3cm}
\subsection{Semantic-aware Contrastive Learning}
To enable the topic representation matrix $E_t$ to capture semantic patterns and group semantically similar words into clusters, we train GCTM-OT using semantic-aware contrastive learning~\cite{DBLP:conf/nips/ChuangRL0J20}.

Concretely, given two batches of inferred document-topic distributions $\Theta=\Theta^a\cup\Theta^b=\{\vec \theta^a_1,\vec \theta^a_2,...,\vec \theta^a_M, \vec \theta^b_1,\vec \theta^b_2,...,\vec \theta^b_M\}$ from augmented documents $X=X^a\cup X^b=\{x_1^a,x_2^a,...,x_M^a,..., x_1^b,x_2^b,...,x_M^b\}$, for each document-topic distribution $\vec \theta^a_i$, we first select the corresponding $\vec \theta^b_i$ to form the positive pair ($\vec \theta_i^a,\vec \theta^b_i$). Then, we select $\vec \theta_{\neg}\in \Theta$ from a semantically irrelevant $x_{\neg}\in X$ to form negative pairs ($\vec \theta^a_i, \vec \theta_{\neg}$). Here, $\vec \theta_{\neg}$ should satisfy the following requirements: 1). $\vec \theta_{\neg}\notin\{\vec \theta^a_i,\vec \theta^b_i\}$. 2). $\cos(\hat{\mathcal{T}}(x_i^a),\hat{\mathcal{T}}(x_{\neg}))<\delta$, where $\hat{\mathcal{T}}(\cdot)$ denotes the document representation obtained by averaging the contextualized word representations produced by the transformer $\mathcal{T}$, $\delta$ is the similarity threshold hyperparameter, set to 0.6 in the experiments.

Thus, the contrastive objective of the augmented document $x^a_i$ could be computed with:
\begin{equation}
    l_i^a\!=\!-\!\log \frac{\textrm{e}^{[\textrm{s}(\vec \theta^a_i,\vec \theta^b_i)/\tau]}}{\Delta\!+\!\!\!\sum\limits_{j=1}^{M}\![\textrm{ind}(\!\vec \theta^a_i\!,\!\vec \theta^a_j)\textrm{e}^{[\textrm{s}(\vec \theta^a_i\!,\vec \theta^a_j)\!/\!\tau\!]}\!+\!\textrm{ind}(\!\vec \theta^a_i\!,\!\vec \theta^b_j)\textrm{e}^{[\textrm{s}(\vec \theta^a_i\!,\vec \theta^b_j)\!/\!\tau\!]}]}
    \label{cs1}
\end{equation}
where $\textrm{s}(\cdot,\cdot)$ is the cosine similarity, $\Delta=\textrm{e}^{[\textrm{s}(\vec \theta^a_i,\vec \theta^b_i)/\tau]}$, and $\tau$ represents the temperature parameter (set to 0.05), the indicator function $\textrm{ind}(\vec \theta_i^u, \vec \theta_j^v)$ is defined as:
\begin{equation}
    \textrm{ind}(\vec\theta^u_i,\vec\theta^v_j)=\left\{  
    \begin{aligned}
        &1, \quad i\neq j \ and \   \cos(\hat{\mathcal{T}}(x_i^u),\hat{\mathcal{T}}(x_j^v))<\delta\\
        &0, \quad \quad \quad \quad otherwise
    \end{aligned}
    \right. 
    \label{cs2}
\end{equation}
where $u,v\!\in\! \{a,b\}$. Overall, the contrastive objective of two batches of augmented documents $X=X^a\cup X^b$ could be computed with:
\begin{equation}
    L_X=\frac{1}{2M}\sum_{i=1}^M(l_i^a+l_i^b)
    \label{cs3}
\end{equation}
where $M$ denotes the batch size.

\vspace{-0.3cm}
\subsection{Goal-oriented Optimal Transport Alignment}
\label{sec:goalrepresentation}

To inject goal information into the learned topics more directly and ensure topic diversity, we identify key goal candidates from the corpus-level goal set $G$ via clustering and use them to guide model training through goal-oriented optimal transport alignment. 

Specifically, given the constructed goal candidates set $G$, we first identify a set of $K$ key goals by grouping them into clusters with:
\begin{equation}
    [C_{1},C_{2},...,C_{K}]=\textrm{KMeans}(\bar{\mathcal{T}}(G))
    \label{kmeans}
\end{equation}
where $\bar{\mathcal{T}}(\cdot)$ denotes the mean of token embeddings produced by $\mathcal{T}$ for goal candidates in $G$. $C_k$ ($k \in \{1, 2, \ldots, K\}$) represents the $k$-th cluster of goal candidates. For each cluster $C_k$, we use its centroid to form the $k$-th goal representation $\vec e_{g}^{k}\in E_g$. Here, $E_g\in \mathbb{R}^{K\times H}$ denotes the goal representation matrix, $H$ is the dimensionality of the transformer representations.

Then, for each augmented document $x^a$, we infer the corresponding document-goal distribution $\vec \lambda^a$ with:
\begin{equation}
    \vec \lambda^a=\textrm{softmax}([\cos(\vec h^a,\vec e^{1}_g),\cos(\vec h^a,\vec e^{2}_g),...,\cos(\vec h^a,\vec e^{K}_g)])
    \label{inferlambda}
\end{equation}
where $\vec h^a$ denotes the contextualized representation of the augmented document $x^a$, $\textrm{softmax}(\cdot)$ and $\cos(\cdot,\cdot)$ denote the softmax normalization and cosine similarity.

Thus, to inject the goal information into the modeling process and influence the learning of the topic representation matrix $E_t$, we align the document-topic distribution $\vec \theta^{a}$ with the document-goal distribution $\vec \lambda^a$ using optimal transport (OT)~\cite{DBLP:journals/ftml/PeyreC19}, where the distance $d_{\hat{\bf{M}}}(\vec \theta^a,\vec \lambda^a)$ is formulated as:
\begin{equation}
    d_{\hat{\bf{M}}}(\vec \theta^a, \vec \lambda^a)=\min\limits_{ {\bf{P}} \in U(\vec \theta^a,\vec \lambda^a)}\langle \bf{P}, \hat{\bf{M}} \rangle \label{eq:ot}
\end{equation}
where $\langle \cdot,\cdot\rangle$ means the Frobenius dot-product, ${\bf{P}}\!\!\in\!\! \mathbb{R}^{K\times K}_{>0}$ is the transport matrix, $U(\vec \theta^a,\vec \lambda^a)$ is the transport polytope of $\vec \theta^a$ and $\vec \lambda^a$, which is the polyhedral set of $K\times K$ matrix: $U(\vec \theta^a,\vec \lambda^a)\!:=\!\{P\in \mathbb{R}_{>0}^{K\times K}|P{\vec {1}}_{K}\!=\!\vec \theta^a, P^{T}{\vec {1}}_{K}\!=\!\vec \lambda^a\}$, and $\vec 1_{K}$ is the $K$-dimensional vector of ones. $\hat{\bf{M}}\!\in\! \mathbb{R}^{K\times K}_{>0}$ is the cost matrix of the transport, where the element $m_{i,j}$ in the $i$-th row and $j$-th column is formed with:
\begin{equation}
m_{i,j}=1-\cos(\vec e^{i}_t,\vec e^{j}_g)
\label{costmatrix}
\end{equation}
where $\cos(\cdot,\cdot)$ denotes cosine similarity, and $m_{i,j}\in [0,2]$, $\vec e_t^i$ and $\vec e_g^j$ denote the $i$-th topic representation and the $j$-th goal representation, respectively. Similarly, we could calculate the OT distance for the augmented document $x^b$ using Eq.~\ref{eq:ot} by feeding in $\vec \theta^{b}$ and $\vec \lambda^b$. As direct computation via Eq.~\ref{eq:ot} is time-consuming for large-scale problems, we follow an entropy-regularized estimator~\cite{DBLP:conf/nips/Cuturi13} and utilize the Sinkhorn iteration~\cite{DBLP:conf/uai/PatriniBFCBWGN19} for the OT distance estimation. 

Overall, given two batches of document-topic distributions $\Theta =\Theta^a\cup \Theta^b$ and the corresponding document-goal distributions $\Lambda=\Lambda^a\cup \Lambda^b$, generated from the augmented documents $X^a\cup X^b$, their optimal transport objective could be formulated as:
\begin{equation}
    OT(\Theta,\Lambda)=\frac{1}{2M}\sum_{i=1}^M[d_{\hat{\bf{M}}}(\vec \theta^a_i, \vec \lambda^a_{i})+d_{\hat{\bf{M}}}(\vec \theta^b_{i}, \vec \lambda^b_{i})] \label{eq:otloss}
\end{equation}
where $M$ denotes the batch size, $\vec \theta^a_{i}$ and $\vec \theta^b_{i}$ are the $i$-th document-topic distributions in $\Theta^a$ and $\Theta^b$, $\vec \lambda^a_{i}$ and $\vec \lambda^b_{i}$ denote the $i$-th document-goal distributions in $\Lambda^a$ and $\Lambda^b$.

\vspace{-0.1cm}
\subsection{Training Objective and Learning Procedure}
To extract goal-relevant, interpretable and diverse topics, our proposed GCTM-OT should take the following factors into account:
\begin{itemize}
    \item Maximizing the similarities between positive pairs~\footnote{Both ($\vec \theta^a$, $\vec \theta^b$) and ($\vec \theta^b$, $\vec \theta^a$).}  and minimizing similarities between negative pairs~\footnote{Both ($\vec \theta^a,\vec \theta_{\neg}$) and ($\vec \theta^b,\vec \theta_{\neg}$).} using semantic-aware contrastive learning, enabling topic representation matrix $E_t$ to capture semantic patterns. 
    \item Matching the inferred document-topic distributions $\Theta=\Theta^a\cup\Theta^b$ to the Dirichlet prior $\textrm{Dir}(\vec \theta'|\vec \alpha)$, ensuring topic interpretability. 
    \item Aligning the inferred document-topic distributions $\vec{\theta}^a$ and $\vec{\theta}^b$ with the document-goal distributions $\vec{\lambda}^a$ and $\vec{\lambda}^b$ using optimal transport, ensuring both semantic relevance to the goals and topic diversity. 
\end{itemize}

Thus, we formulate the training objective of GCTM-OT as:
\begin{equation}
    \mathcal{L}=\mathcal{L}_{c}+\eta \mathcal{L}_{PM}+\zeta \mathcal{L}_{OT} \label{eq:overall}
\end{equation}
where $\mathcal{L}_c$ denotes the contrastive learning objective, computed using Eq.~\ref{cs3}. $\mathcal{L}_{PM}$ and $\mathcal{L}_{OT}$ represent the prior matching and optimal transport objectives, calculated using Eq.~\ref{mmd} and Eq.~\ref{eq:otloss}. $\eta$ and $\zeta$ are coefficient hyperparameters, we set them to 1.0 in our experiments. The detailed training procedure of GCTM-OT is shown in Algorithm~\ref{alg:1}. {\color{black}In our experiment, batch size $M$ is set to 32, learning rate $\alpha_1$ is set to 2e-3, the optimal transport objective $\mathcal{L}_{OT}$ is computed using the GeomLoss~\footnote{\url{https://www.kernel-operations.io/geomloss/}} library with default configuration. GCTM-OT is optimized by Adam~\cite{DBLP:journals/corr/KingmaB14} optimizer. }

\vspace{-0.2cm}

\begin{algorithm}[H]
\centering
\footnotesize
\caption{{\color{black}The training procedure of GCTM-OT.}}
\begin{algorithmic}[1]
\Require  
       Corpus $D$, human's goal $\mathcal{H}_g$, topic number $K$, batch size $M$, learning rate $\alpha_1$.

\For{each document $x$ in $D$}
    \State Generate the document-specific goal candidates set $G_x=\mathcal{P}_s(x)$. 
    \State Add goal candidates in $G_x$ to the corpus-level goal set $G$.
\EndFor

\State Randomly initialize the topic representation matrix $E_t$.
\State Obtain goal representation matrix $E_g$ via Eq.\ref{kmeans}. 
\State Calculate the cost matrix $\hat{\bf{M}}$ of optimal transport using Eq.\ref{costmatrix}. 
\For{each batch of documents $X$ in $D$}
\For{each document $x$ in $X$}
    \State Generate the augmented document pair via $x^a=\mathcal{P}_a(x),\ x^b=\mathcal{P}_a(x)$. 
    \State Construct document representations $\vec h^a$ and $\vec h^b$ for $x^a$ and $x^b$ via Eq.~\ref{c4}. 
    \State Infer document-topic distributions  $\vec\theta^a$ and $\vec\theta^b$ for $x^a$ and $x^b$ via Eq.\ref{eq:inference}.
    \State Infer document-goal distributions $\vec\lambda^a$ and $\vec\lambda^b$ for $x^a$ and $x^b$ via Eq.\ref{inferlambda}.
\EndFor
\State  Calculate  contrastive objective $\mathcal{L}_c$ via Eq.~\ref{cs3}. 
\State  Draw a set of $2M$ random samples $\Theta'$ from $Dir(\vec \theta'|\vec \alpha)$. 
\State  Calculate  the prior matching objective $\mathcal{L}_{PM}$ via Eq.~\ref{mmd}. 
\State Calculate the OT objective $\mathcal{L}_{OT}$ in Eq.~\ref{eq:ot} through Sinkhorn Iterations~\cite{DBLP:journals/ftml/PeyreC19}. 
\State Calculate the overall training objective $\mathcal{L}$ via Eq.~\ref{eq:overall}. 
\State Update topic representation matrix $E_t$ with gradient descent.

\EndFor
\Ensure 
       {\color{black}Learned topic representation matrix $E_t$.}

\end{algorithmic}
\label{alg:1}
\end{algorithm}

\subsection{Topic Extraction}
Leveraging the learned topic representation matrix $E_t$ and the constructed goal representation matrix $E_g$, we could extract goal-relevant topics represented by semantically coherent keywords and interpretable topic summaries.

Specifically, for the $v$-th $(v\in\{1,2,...,V\})$ word $w_v$ in the vocabulary, we first use the transformer $\mathcal{T}$ to generate the contextualized word representations $\vec e_{v}^{n'}$ $(n'\in\{1,2,...,N'_v\})$ of all its appearances and collect them into the word representation matrix $E_{v}\in\mathbb{R}^{H\times N'_{v}}$. Here, $N'_v$ denotes the number of times $w_v$ appears in the corpus, $H$ is the dimensionality of word representations. We then compute the semantic correlations $\vec c_v$ between word $w_v$ and topics using:
\begin{align}
    \vec c_v^{n'}=\textrm{softmax}([\cos(\vec e_{v}^{n'},\vec e_t^1)&,\cos(\vec e_{v}^{n'},\vec e_t^2),...,\cos(\vec e_{v}^{n'},\vec e_t^K)])\\
    \vec c_v=\frac{1}{N'_v}&\sum\nolimits_{n'=1}^{N'_v}\vec c_v^{n'} 
\end{align}
where $\vec{c}_v^{n'}$ denotes the local semantic correlations between the word $w_v$ and topics in its $n'$-th appearance, and $\vec e_{t}^k$ denotes the topic representation of the $k$-th topic.

Similarly, we construct a correlation matrix $\hat{\mathcal{C}} \in \mathbb{R}^{K \times V}$, which stores the semantic correlations between topics and words in the vocabulary. The topic-word distribution $\vec{\phi}_k$ of the $k$-th topic ($k \in \{1, 2, \ldots, K\}$) is then computed with:
\begin{equation}
    \vec \phi_k=\textrm{norm}(\vec c_{k,\cdot})
\end{equation}
where $\vec c_{k,\cdot}$ denotes the $k$-th row of $\hat{\mathcal{C}}$, and $\textrm{norm}(\cdot)$ denotes the normalization function. Moreover, to enhance topic interpretability, GCTM-OT generates a human-readable summary $t_k^s$ for the $k$-th topic using:
\begin{equation}
    t_k^s=\textrm{arg}\max_{\hat{g}\in C_k} (\cos(\bar{\mathcal{T}}(\hat{g}),\vec e_g^k))
\end{equation}
where $\hat{g}$ denotes a goal candidate in the $k$-th cluster $C_k$, $\bar{\mathcal{T}}(\hat{g})$ is the mean of token embeddings produced by transformer $\mathcal{T}$ for $\hat{g}$, and $\vec e_g^k$ is the $k$-th goal representation.

\section{Experiments}
We first introduce the experimental setup, including descriptions of the datasets, baselines, metrics and implementation details. We then present the topic evaluation results along with the corresponding analysis, followed by hyperparameter analysis and ablation studies.

\subsection{Experimental Setup}

\subsubsection{Datasets}
We evaluate the performance of GCTM-OT for \emph{Human-TM} task on three subreddit datasets: \emph{`Bothering'}~\footnote{\url{https://www.reddit.com/r/whatsbotheringyou/}}, \emph{`TeslaModel3'}~\footnote{\url{https://www.reddit.com/r/TeslaModel3/}} and \emph{`AskAcademia'}~\footnote{\url{https://www.reddit.com/r/AskAcademia/}}. Specifically, the \emph{`Whatsbotheringyou'} dataset is a collection of posts about personal concerns, such as `breakup struggle' and `health issue', and is abbreviated as `Bothering' in our experiments. \emph{`TeslaModel3'} dataset contains user-generated posts covering various aspects of this electric vehicle, such as `charging speed' and `autopilot'. \emph{`AskAcademia'} dataset contains posts about academic life, including topics like `work–life balance' and `research challenges'. All datasets are sourced from Pushshift’s Reddit data dumps\footnote{\url{https://www.reddit.com/r/pushshift/comments/1akrhg3/separate_dump_files_for_the_top_40k_subreddits/}}. For each dataset, to ensure data quality, we remove duplicate posts with the same \emph{post\_id}, low-engagement posts with scores below 10, and irrelevant posts tagged by \emph{moderators}. We then perform a series of preprocessing steps on each selected post, including spell checking, stemming and stopword removal, using the SpaCy\footnote{\url{https://spacy.io/}} library. The statistics of the processed datasets are presented in Table~\ref{tab:Dataset}.
\begin{table}[htbp]
    \centering
    \vspace{-0.3cm}
    \resizebox{0.28\textwidth}{!}{
    \begin{tabular}{c|ccc}
        \toprule
        \textbf{Datasets}&\textbf{\#Doc}&\textbf{\#Vocab}&\textbf{ Avg\_Len}  \\
        \midrule
         Bothering&1,483&7,700&56.48\\
         TeslaModel3&1,991&5,634&32.82\\
         AskAcademia &3,348&8,122&36.16\\
         \bottomrule
    \end{tabular}
}
    \caption{The statistics of processed datasets.}
    \vspace{-0.5cm}
    \label{tab:Dataset}
\end{table}


\subsubsection{Baselines}

We choose the following approaches as baselines:

\begin{itemize}
    \item \underline{\textbf{LDA}}~\citep{DBLP:journals/jmlr/BleiNJ03} is a probabilistic topic model, which models documents with a mixture of topics. We use the Mallet~\footnote{\url{https://github.com/mimno/Mallet}} implementation in the experiments. 
    \item \underline{\textbf{BAT}}~\cite{DBLP:conf/acl/WangHZHXYX20} is a neural topic modeling approach based on bidirectional adversarial training. We implement the approach following the default parameter configurations. 
    \item \underline{\textbf{CTMNeg}}~\cite{DBLP:conf/icon-nlp/AdhyaLSD22} is a neural topic model based on VAE and negative sampling, we use the official implementation~\footnote{\url{https://github.com/adhyasuman/ctmneg}}.
    \item \underline{\textbf{vONT}}~\cite{DBLP:conf/acl/XuJRIZ23} is a neural topic model, which models topics with a mixture of von-Mises Fisher distributions. We use the official implementation~\footnote{\url{https://github.com/xuweijieshuai/Neural-Topic-Modeling-vmf}}. 
    \item \underline{\textbf{CWTM}}~\cite{DBLP:conf/coling/00110P24} is a topic model based on the contextualized representations, we use the official implementation~\footnote{\url{DBLP:conf/coling/00110P24}}. 
    \item \underline{\textbf{DisCTM}}\cite{DBLP:conf/wsdm/WangLWCYH25} is a contextualized neural topic model for short text modeling. We implement the approach following the default parameter configurations. 
    \item \underline{\textbf{HiCOT}}\cite{vuong2025hicot} is a neural topic model based on optimal transport, we use the official implementation~\footnote{\url{https://github.com/HoangTran223/HiCOT}}. 
    \item \underline{\textbf{CAST}}\cite{DBLP:conf/naacl/MaXYVHLN25} is a topic model that leverages word embeddings to remove functional words for topic extraction\footnote{\url{https://github.com/yananma1029/CAST}}. 
    \item \underline{\textbf{LLM-TE}}~\cite{DBLP:conf/coling/MuDBS24} is a topic mining approach based on prompt engineering, we use the official implementation~\footnote{\url{https://github.com/GateNLP/LLMs-for-Topic-Modeling}}. 
    \item \underline{\textbf{LLM-ITL}}~\cite{DBLP:journals/corr/abs-2411-08534} is a neural topic model that incorporates an LLM-based topic refinement mechanism~\footnote{\url{https://github.com/Xiaohao-Yang/LLM-ITL}}. 

\end{itemize}

\subsubsection{Evaluation Metrics} \emph{Human-TM}'s performance is evaluated using topic coherence and diversity metrics. In addition, we propose three novel metrics to assess the relevance between human-provided goals and the extracted topics.

To measure topic interpretability, we follow~\cite{DBLP:conf/wsdm/RoderBH15} and employ four widely used topic coherence metrics: $C_P$, $C_A$, $NPMI$ and $UCI$. All coherence values are computed using the Palmetto~\footnote{\url{https://github.com/dice-group/Palmetto}} library, with higher values indicating greater interpretability.

For topic diversity evaluation, we follow~\cite{DBLP:journals/tnn/WangZHZ25} and use the Unique Term ($UT$) metric, defined as:
\begin{equation}
    UT=\#N_u/(10\times K)
\end{equation}
where $\#N_u$ denotes the number of unique words among the topic words, and $K$ is the number of topics.

\begin{figure*}
    \centering
    \vspace{-0.2cm}
    \includegraphics[width=0.9\linewidth]{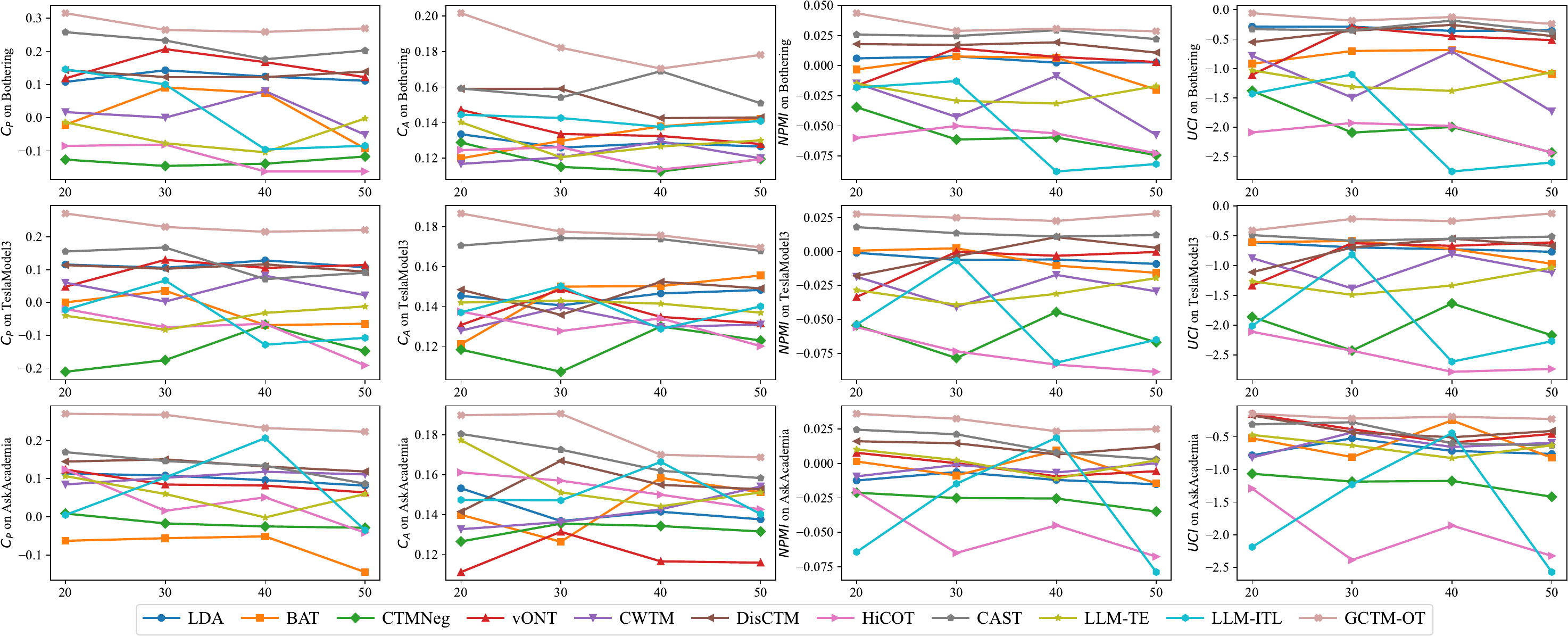}
    \vspace{-0.2cm}
    \caption{The comparison of topic coherence values vs. different topic number settings [20, 30, 40, 50].}
    \label{fig:mainresult}
\end{figure*}

  \begin{table*}[htbp]
    \centering
    \resizebox{0.9\textwidth}{!}{
  \begin{tabular}{l|cccc|cccc|cccc}
  \toprule
  \textbf{Dataset} & \multicolumn{4}{c|}{\textbf{Bothering}}        & \multicolumn{4}{c|}{\textbf{TeslaModel3}}      & \multicolumn{4}{c}{\textbf{AskAcademia}} \\
  \midrule
  \textbf{Model} & $\bm{C_P}$ & $\bm{C_A}$ & $\bm{NPMI}$ & $\bm{UCI}$  & $\bm{C_P}$ & $\bm{C_A}$ & $\bm{NPMI}$ & $\bm{UCI}$   & $\bm{C_P}$ & $\bm{C_A}$ & $\bm{NPMI}$ & $\bm{UCI}$ \\
  \midrule
  LDA   & 0.1214  & 0.1286  & 0.0047  & -0.3226    & 0.1143  & 0.1451  & -0.0056  & -0.6976    & 0.0997  & 0.1422  & -0.0114  & -0.6964   \\
  BAT   & 0.0127  & 0.1324  & -0.0022  & -0.8499   & -0.0246  & 0.1441  & -0.0058  & -0.7193    & -0.0786  & 0.1439  & -0.0031  & -0.6011    \\
  CTMNeg & -0.1319  & 0.1189  & -0.0575  & -1.9725   & -0.1510  & 0.1195  & -0.0611  & -2.0211    & -0.0156  & 0.1319  & -0.0267  & -1.2121    \\
  vONT  & 0.1535  & 0.1353  & 0.0020  & -0.5920    & 0.0995  & 0.1363  & -0.0093  & -0.8079    & 0.0883  & 0.1187  & -0.0018  & -0.3956    \\
  CWTM  & 0.0175  & 0.1132  & -0.0107  & -0.6486    & 0.0417  & 0.1320  & -0.0265  & -1.0433    & 0.1039  & 0.1414  & -0.0042  & -0.6252   \\
  DisCTM & 0.1316  & 0.1509  & 0.0163  & -0.4035    & 0.1067  & 0.1464  & -0.0020  & -0.7551    & 0.1362  & 0.1540  & 0.0124  & -0.3812    \\
  HiCOT & -0.1226  & 0.1209  & -0.0599  & -2.1051    & -0.0881  & 0.1298  & -0.0753  & -2.5121    & 0.0372  & 0.1527  & -0.0494  & -1.9659   \\
  CAST  & 0.2097  & 0.1584  & 0.0254  & -0.2828    & 0.1212  & 0.1716  & 0.0136  & -0.5318   & 0.1337  & 0.1683  & 0.0142  & -0.4551    \\
  LLM-TE & 0.0164  & 0.1414  & -0.0502  & -1.9701    & -0.0480  & 0.1389  & -0.0519  & -1.9261    & 0.0696  & 0.1502  & -0.0348  & -1.6095   \\
  LLM-ITL & -0.0493  & 0.1293  & -0.0232  & -1.1970    & -0.0419  & 0.1408  & -0.0296  & -1.2793   & 0.0561  & 0.1559  & 0.0010  & -0.6401   \\
  GCTM-OT & \textbf{0.3326} & \textbf{0.1916} & \textbf{0.0419} & \textbf{-0.1020}  & \textbf{0.3031} & \textbf{0.2011} & \textbf{0.0377} & \textbf{-0.1889}  & \textbf{0.2990} & \textbf{0.1946} & \textbf{0.0405} & \textbf{-0.1173} \\
  \bottomrule
  \end{tabular}%

}
    \caption{Average topic coherence values with four topic number settings [20, 30, 40, 50].}
\vspace{-0.3cm}
    \label{tab:mainresult}
  \end{table*}%

Furthermore, to assess the relevance between the human-provided goal $\mathcal{H}_g$ and the extracted topics, we propose three novel metrics: Goal Similarity ($GS$), Goal-relevant Topic Rate ($GTR$) and Goal Coverage Rate ($GCR$), all computed based on OpenAI embeddings~\footnote{\url{https://platform.openai.com/docs/guides/embeddings}}. 

Concretely, given a corpus $D$ and the human-provided goal $\mathcal{H}g$, we first identify a set of $N_g$ ground-truth goal candidates $\hat{G}=\{\hat{g}{1}, \hat{g}{2},\dots,\hat{g}{N_g}\}$ through clustering and human annotation. To measure the semantic relevance between the extracted topics $T=\{t_1,t_2,\dots,t_K\}$ and the ground-truth goal candidates $\hat{G}$, we define the Goal Similarity ($GS$) metric with:
\begin{equation}
    GS(T,\hat{G})=\frac{1}{K}\sum\nolimits_{k=1}^K \max_{\hat{g}\in \hat{G}}\cos(\vec e_{t_k},\vec e_{\hat{g}})\label{eq:gs}
\end{equation}
where $\vec e_{t_k}$ and $\vec e_{\hat{g}}$ denote the OpenAI embeddings of the $k$-th topic $t_k$ and the goal candidate $\hat{g}$, respectively. On the other hand, to measure the proportion of topics that are relevant to the goals, we define the Goal-relevant Topic Rate ($GTR$) metric as:
\begin{equation}
    GTR(T,\hat{G})=\frac{1}{K}\sum\nolimits_{k=1}^K\max (0,\textrm{sgn}(\max_{\hat{g}\in \hat{G}}\cos(\vec e_{t_k},\vec e_{\hat{g}})-\sigma_1))\label{eq:gtr}
\end{equation}
where $\cos(\vec e_{t_k},\vec e_{\hat{g}})$ means the cosine similarity between OpenAI embeddings of the $k$-th topic and the goal candidate $\hat{g}$, $\textrm{sgn}(\cdot)$ denotes the sign function, and $\sigma_1$ is the threshold hyperparameter. Similarly, to measure the proportion of goal candidates in $\hat{G}$ that the extracted topics can discover, we define the Goal Coverage Rate ($GCR$) metric as:
\begin{equation}
    GCR(T,\hat{G})=\frac{1}{N_{g}}\sum\nolimits_{i=1}^{N_g}\max (0,\textrm{sgn}(\max_{t\in T}\cos(\vec e_{\hat{g_i}},\vec e_{t})-\sigma_2))\label{eq:gcr}
\end{equation}
where $\cos(\vec e_{\hat{g_i}},\vec e_{t})$ is the cosine similarity between OpenAI embeddings of the $i$-th goal candidate $\hat{g}_{i}$ and topic $t$, and $\sigma_2$ denotes the threshold hyperparameter.

\subsubsection{Implementation Details} {\color{black}In our experiments, we use the MPNet~\cite{DBLP:conf/nips/Song0QLL20} ($\mathcal{T}$) to generate contextualized representations. For LLM-based prompting (goal summarization prompt $\mathcal{P}_s$ and text augmentation prompt $\mathcal{P}_a$), we employ the GPT-3.5-turbo. For three datasets, the analytical objectives (goal) are defined with: \emph{``What's the thing that's bothering you?''}, \emph{``What is the specific aspect of TeslaModel3?''} and \emph{``What is the specific concern about academia?''}, respectively.} All experiments are conducted on a machine running Ubuntu, equipped with an NVIDIA RTX 4090 GPU (CUDA 12.2) and 64GB RAM. 


\begin{figure*}[htbp]
\centering
\begin{minipage}{0.6\textwidth}
    \centering
    \resizebox{\textwidth}{!}{
    \begin{tabular}{l|cccc|cccc|cccc}
\toprule
\multicolumn{1}{c|}{\multirow{2}[4]{*}{\textbf{Model}}} & \multicolumn{4}{c|}{\textbf{Bothering}} & \multicolumn{4}{c|}{\textbf{TeslaModel3}} & \multicolumn{4}{c}{\textbf{AskAcademia}} \\
\cmidrule{2-13}        & 20    & 30    & 40    & 50    & 20    & 30    & 40    & 50    & 20    & 30    & 40    & 50 \\
\midrule
LDA   & 0.6950 & 0.6000 & 0.5750 & 0.5620 & 0.6150 & 0.5867 & 0.5600 & 0.5260 & 0.6650 & 0.6333 & 0.5675 & 0.5240 \\
BAT   & 0.5150 & 0.6067 & 0.5525 & 0.6480 & 0.7050 & 0.7633 & 0.6675 & 0.6680 & 0.6550 & 0.5767 & 0.5525 & 0.5160 \\
CTMNeg & 0.7250 & 0.6933 & 0.6800 & 0.6120 & 0.7050 & 0.7133 & 0.6725 & 0.6500 & 0.7650 & 0.7033 & 0.6950 & 0.6140 \\
vONT  & 0.7700 & 0.6600 & 0.6350 & 0.6320 & 0.6600 & 0.5267 & 0.5325 & 0.5100 & 0.8750 & 0.7967 & 0.6750 & 0.6180 \\
CWTM  & 0.4850 & 0.3767 & 0.4000 & 0.4140 & 0.6400 & 0.5133 & 0.4050 & 0.5200 & 0.7000 & 0.6533 & 0.6125 & 0.5200 \\
DisCTM & 0.7100 & 0.6800 & 0.5600 & 0.5120 & 0.8000 & 0.6567 & 0.6200 & 0.5620 & 0.8650 & 0.8067 & 0.7400 & 0.6720 \\
HiCOT & 0.7400 & 0.6433 & 0.5300 & 0.6000 & 0.6950 & 0.6600 & 0.6050 & 0.5400 & 0.6800 & 0.7433 & 0.5300 & 0.5260 \\
CAST  & 0.6300 & 0.6900 & 0.5350 & 0.5600 & 0.8300 & 0.6800 & 0.6375 & 0.5880 & 0.8250 & 0.7100 & 0.6975 & 0.6000 \\
LLM-TE & 0.6750 & 0.4167 & 0.7875 & 0.7460 & 0.8000 & 0.3467 & 0.7800 & 0.6860 & 0.7000 & 0.7267 & 0.5375 & 0.7740 \\
LLM-ITL & 0.7000 & 0.6100 & 0.5950 & 0.5520 & 0.6500 & 0.5900 & 0.5675 & 0.5420 & 0.7150 & 0.5933 & 0.6425 & 0.5920 \\
GCTM-OT & \textbf{0.9800} & \textbf{0.9200} & \textbf{0.8175} & \textbf{0.7400} & \textbf{0.9750} & \textbf{0.9133} & \textbf{0.8125} & \textbf{0.7260} & \textbf{1.0000} & \textbf{0.9400} & \textbf{0.9000} & \textbf{0.8380} \\
\bottomrule
\end{tabular}
    }
    \captionof{table}{Comparison on $UT$ metric across different topic number settings.}
    \vspace{1em}
    \label{tab:ut}
    \vspace{-0.4cm}
    \resizebox{\textwidth}{!}{
    \begin{tabular}{c|ccccccccccc}
  \toprule
  \textbf{Dataset} & \textbf{LDA} & \textbf{BAT} & \textbf{CTMNeg} & \textbf{vONT} & \textbf{CWTM} & \textbf{DisCTM} & \textbf{HiCOT} & \textbf{CAST} & \textbf{LLM-TE} & \textbf{LLM-ITL} & \textbf{GCTM-OT} \\
  \midrule
  \textbf{Bothering} & 0.3611  & 0.3386  & 0.3654  & 0.3593  & 0.3557  & 0.3557  & 0.3356  & 0.3845  & 0.3616  & 0.3653  & \textbf{0.4253}  \\
  \textbf{TeslaModel3} & 0.4607  & 0.3496  & 0.4131  & 0.4367  & 0.4188  & 0.4110  & 0.4051  & 0.4696  & 0.4689  & 0.4736  & \textbf{0.5100}  \\
  \textbf{AskAcademia} & 0.3890  & 0.3248  & 0.3933  & 0.3789  & 0.3635  & 0.3915  & 0.4014  & 0.3943  & 0.3971  & 0.4104  & \textbf{0.4530}  \\
  \bottomrule
  \end{tabular}%
  }
    \captionof{table}{Comparison results on Goal Similarity ($GS$) metric.}
    \label{tab:mms_results}
\end{minipage}
\hfill
\begin{minipage}{0.37\textwidth}
    \centering
    \vspace{-0.6cm}
    \includegraphics[width=\linewidth]{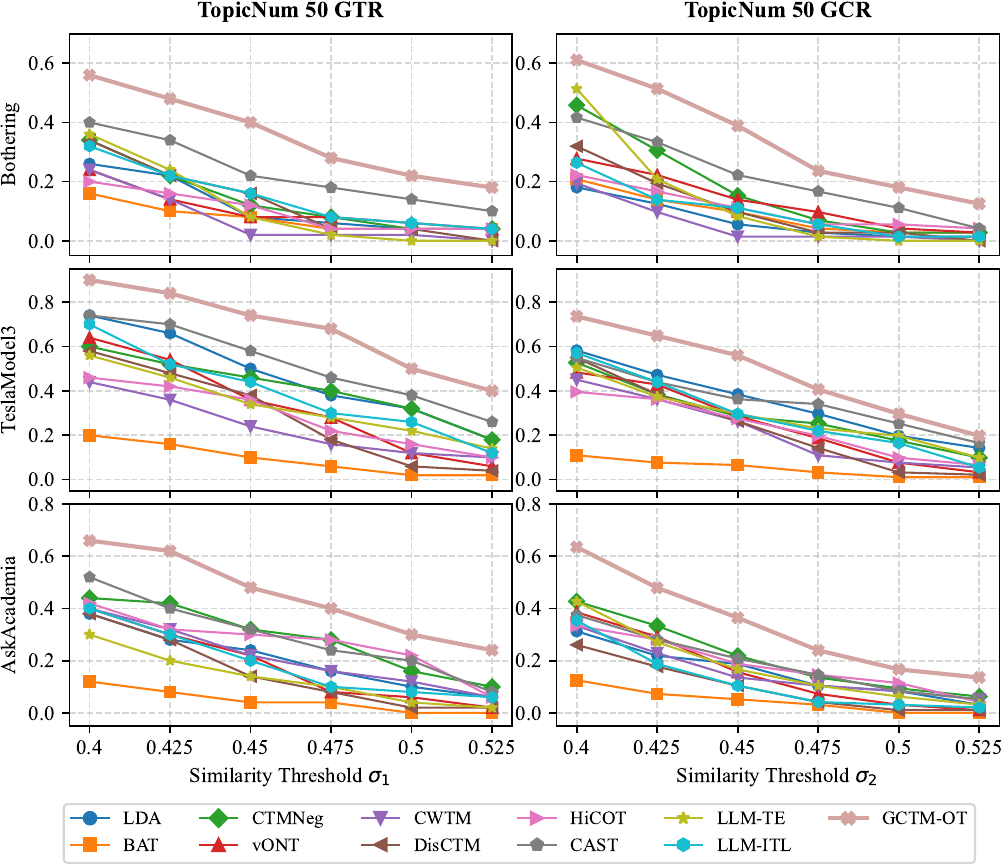}
    \vspace{-0.8cm}
    \caption{Comparision on $GTR$ and $GCR$ metrics.}
    \label{fig:GTR}
\end{minipage}
\vspace{-0.4cm}
\end{figure*}

\vspace{-0.1cm}
\subsection{Topic Evaluation and Results Analysis}

In this section, we conduct extensive experiments on the \emph{`Bothering'}, \emph{`TeslaModel3'} and \emph{`AskAcademia'} datasets, and present the results along with an analysis of topic coherence, diversity and goal-relevance metrics. For each dataset, we run the model with four different topic number configurations [20, 30, 40, 50], and we select the top 10 words according to the topic-word probabilities to represent the meaning of each topic.

\subsubsection{Topic Coherence and Diversity Evaluation} To assess the topic interpretability and diversity, we perform comparisons on topic coherence ($C_P$, $C_A$, $NPMI$ and $UCI$) and diversity ($UT$) metrics.

\textbf{Coherence:} Table~\ref{tab:mainresult} lists the average topic coherence values for the $C_P$, $C_A$, $NPMI$ and $UCI$ metrics across the three datasets.  Each value is obtained by averaging the coherence scores from different topic-number configurations. For example, GPTM’s $NPMI$ value (0.0405) on the \emph{`AskAcademia'} dataset is the mean value of 0.04779 (20 topics), 0.04328 (30 topics), 0.03338 (40 topics) and 0.03751 (50 topics). To examine whether GCTM-OT is sensitive to variations in the topic-number setting, we also present a comparison of topic coherence vs. topic number in Figure~\ref{fig:mainresult}.  The statistics show that GCTM-OT outperforms state-of-the-art baselines across the three datasets under different topic-number settings. These improvements may be attributed to the following aspects: 1). Modeling topics with the Dirichlet prior helps improve topic quality. 2). Incorporating goal information into the text representation process facilitates the generation of meaningful text representations. 3). Incorporating contextualized word representations into topic word extraction is crucial for mining coherent topics.




\textbf{Diversity:} To compare topic diversity, we report the $UT$ values in Table~\ref{tab:ut}, with the optimal results highlighted in bold. The statistics show that GCTM-OT produces more diverse topics than state-of-the-art baselines across the three datasets under different topic-number configurations. The superior performance of GCTM-OT on the $UT$ metric may be attributed to the following two factors: 1). Semantic-aware contrastive learning tends to separate documents with different semantics into distinct clusters, facilitating the discovery of diverse patterns. 2). The goal-oriented optimal transport mechanism aligns topics with different goals, thereby enhancing topic diversity.

\subsubsection{Goal Annotation and Relevence Evluation} To assess the relevance between goal $\mathcal{H}_g$ and extracted topics $T$, we should first identify a set of $N_g$ ground-truth goal candidates $\hat{G}=\{\hat{g_1}, \hat{g_2},...,\hat{g_{N_g}}\}$.


\textbf{Goal Annotation:} For each dataset, we first perform clustering on its corpus-level goal candidate set $G$ and obtain a set of $N_c$~\footnote{$N_c$ is set to 100 in our evaluation.} goal candidate clusters $C_g=\{C_1,C_2,...,C_{N_{c}}\}$. We then recruit two experts (human or LLM) to identify if the $i$-th ($i\in\{1,2,...,N_c\}$) cluster $C_i$ is a reasonable candidate for $\mathcal{H}_{g}$. In our experiments, we employ the Jaccard Similarity Coefficient ($JSC$)~\cite{DBLP:journals/prl/Kosub19}, Simple Matching Coefficient ($SMC$)~\cite{goodall1967distribution} and the Sorensen-Dice Coefficient ($SDC$)~\cite{carass2020evaluating} to measure the agreement between the two experts’ annotation results, with the coefficient values listed in Table~\ref{tab:agreement}. Only clusters labeled as true by both human experts are added to the ground-truth goal candidate set $\hat{G}$, resulting in 72, 91 and 96 ground-truth goal candidates for \emph{`Bothering'}, \emph{`TeslaModel3'} and \emph{`AskAcademia'}.

\begin{table}[!t]
  \centering
  \resizebox{0.4\textwidth}{!}{
  \begin{tabular}{l|ccc|ccc}
    \toprule
    \multirow{2}{*}{\textbf{Dataset}} & \multicolumn{3}{c|}{\textbf{Human vs. Human}} & \multicolumn{3}{c}{\textbf{Human vs. LLM}} \\
    \cmidrule(lr){2-4} \cmidrule(lr){5-7}
    & $\textbf{$\bm{JS}$}$ & $\textbf{$\bm{SMC}$}$ & $\textbf{$\bm{DSC}$}$ & $\textbf{$\bm{JS}$}$ & $\textbf{$\bm{SMC}$}$ & $\textbf{$\bm{DSC}$}$ \\
    \midrule
    Bothering & 0.7579  & 0.7700  & 0.8623  & 0.7738  & 0.7750  & 0.8722  \\
    TeslaModel3 & 0.9192  & 0.9200  & 0.9579  & 0.8800  & 0.8800  & 0.9357  \\
    AskAcademia & 0.9600  & 0.9600  & 0.9796  & 0.9196  & 0.9200  & 0.9581  \\
    \bottomrule
  \end{tabular}
}
  \caption{Coefficients between two experts' annotation results.}
\vspace{-0.7cm}
  \label{tab:agreement}
\end{table}


\textbf{Goal-relevance Evaluation:} To assess the semantic relevance between the topics and the annotated goal candidate set $\hat{G}$, we calculate the Goal Similarity ($GS$) values using  Eq.~\ref{eq:gs} and present the results in Table~\ref{tab:mms_results}. Each value is the mean $GS$ value across four topic number setting, and $GS$ value is obtained by averaging the maximum semantic similarity between each topic $t_k \in T$ and the goal candidates in $\hat{G}$. The statistics show that our proposed GCTM-OT consistently generates topics with higher semantic relevance to human-provided goals than the baselines, across all three datasets and under various topic number settings.

\begin{table*}[!t]
  \centering
  \vspace{-0.2cm}
\begin{tabular}{c|c|l}
\toprule
\multicolumn{1}{c|}{\textbf{Model}} & \textbf{Topic Summary} & \multicolumn{1}{c}{\textbf{Topic Words}} \\
\midrule
\multirow{3}{*}{LDA} & --- & \emph{completely, walk}, girlfriend, \emph{hang}, answer, \emph{finally, happy,} honestly \emph{text} couple \\
 & --- & pain, issue, tired, doctor, knee, \emph{sister}, sleep, \emph{clean, house,} tear \\
 & --- & money, college, afford, bill, rent, apply, house, plan, \emph{student, boss} \\
\midrule
\multirow{3}{*}{LLM-ITL} & --- & loved, relationship, past, old, \emph{text, feel,} love, emotional, \emph{life, talk} \\
 & --- & doctor, hospital, mental, cancer, symptom, helping, \emph{degree, currently, training,} surgery
 \\
 & --- & money, work, need, \emph{wanted, open,} using, \emph{month,} job, \emph{know, asking} \\
\midrule
\multirow{3}{*}{GCTM-OT} & Breakup Struggles & cheat, breakup, girlfriend, betray, ring, break, divorce, relationship, partner, devastate \\
 & Health Issue & lung, surgery, cancer, knee, blood, diagnosis, medical, condition, doctor, injury \\
 & Financial Strain & financially, financial, expense, afford, money, payment, debt, income, bill, expensive \\
\bottomrule
\end{tabular}
  \caption{Topic Examples extracted from the \emph{`Bothering'} dataset, \emph{italic} means out-of-topic words.}
  \vspace{-0.3cm}
  \label{tab:example}

\end{table*}

Similarly, to assess the proportion of topics that are relevant to goals and the proportion of goals in $\hat{G}$ that extracted topics could discover, we calculate the Goal-relevant Topic Rate ($GTR$) and the Goal Coverage Rate ($GCR$) values using Eq.~\ref{eq:gtr} and Eq.~\ref{eq:gcr}. For evaluation, we set the threshold hyperparameters ($\sigma_1$ and $\sigma_2$) to six different values: [0.4, 0.425, 0.45, 0.475, 0.5, 0.525], columns mean the metrics, and rows denote datasets. The comparison results for $GTR$ and $GCR$ in Figure~\ref{fig:GTR} show that GCTM-OT generates more goal-oriented topics than the baselines and that these topics cover a broader range of goal candidates in $\hat{G}$.

The superior performance of GCTM-OT on the $GS$, $GTR$ and $GCR$ metrics may be attributed to the following factors: 1). Goal-oriented text representation mechanism injects goal information into text representations and topics. 2). Goal-oriented optimal transport mechanism aligns the trainable topic representation matrix $E_t$ with the LLM-generated goal representation matrix $E_g$ via optimal transport, thereby facilitating goal-relevant topic generation.


\begin{table}[!t]
  \centering
  \resizebox{0.46\textwidth}{!}{
  \begin{tabular}{c|c|ccccc}
  \toprule
  \textbf{Parameter} &  \textbf{Setting} & $\bm{C_P}$    & \textbf{$\bm{C_A}$}    & \textbf{$\bm{NPMI}$}  &  \textbf{$\bm{UCI}$}  &  \textbf{$\bm{UT}$} \\
  \midrule
  \multicolumn{2}{c|}{CAST} & \multicolumn{1}{c}{0.2575} & 0.1592  & 0.0258  & -0.3314  & 0.6300  \\
  \midrule
  \multirow{5}[2]{*}{$\delta$} & 0.2   & 0.3161  & 0.2001  & 0.0408  & -0.1491  & 0.9700  \\
        & 0.4   & 0.3431  & 0.1891  & 0.0414  & -0.2578  & 0.9650  \\
        & 0.6   & \textbf{0.3518} & \textbf{0.2084} & \textbf{0.0524} & \textbf{0.0568} & 0.9800  \\
        & 0.8   & 0.3371  & 0.1962  & 0.0416  & -0.1541  &  \textbf{0.9850} \\
        & 1     & 0.2959  & 0.1957  & 0.0329  & -0.2770  & 0.9450  \\
  \midrule
  \multirow{5}[2]{*}{$\eta$} &0.6 &  0.3251  & \textbf{0.2134} & 0.0387  & -0.2748  & \textbf{0.9800}  \\
        & 0.8   & \textbf{0.3593} & 0.2108 & 0.0491  & -0.1147  & 0.9600  \\
        & 1     & 0.3518  & 0.2084  & \textbf{0.0524} & \textbf{0.0568} & \textbf{0.9800} \\
        & 1.2   & 0.3492  & 0.2023  & 0.0406  & -0.2481  & 0.9750  \\
        & 1.4   & 0.3472  & 0.1942  & 0.0433  & -0.1644  & \textbf{0.9800} \\
  \midrule
  \multirow{5}[2]{*}{$\zeta$} & 0.6   & 0.3318  & 0.2057  & 0.0436  & -0.1256  & 0.9650  \\
        & 0.8   & 0.3274  & 0.2037  & 0.0396  & -0.2125  & 0.9700  \\
        & 1     & \textbf{0.3518} & 0.2084  & \textbf{0.0524} & \textbf{0.0568} & \textbf{0.9800} \\
        & 1.2   & 0.3293  & 0.1917  & 0.0381  & -0.2889  & 0.9450  \\
        & 1.4   & 0.3348  & \textbf{0.2125} & 0.0448  & -0.1573  & 0.9550  \\
  \bottomrule
  \end{tabular}%
}

  \caption{Comparison results of hyperparameter analysis.}
\vspace{-0.4cm}
  \label{tab:Sensitivity_Analysis}%
\end{table}%

To provide an intuitive comparison of topic interpretability, we present three example topics—\emph{`Breakup Struggle'}, \emph{`Health Issue'} and \emph{`Financial Strain'}—in Table~\ref{tab:example}. We observe that GCTM-OT generates interpretable, goal-oriented topics accompanied by automatically produced, human-readable topic summaries. The symbol `$-$' indicates that topic summaries are not available for LDA and LLM-ITL.


\subsection{Hyperparameter Analysis}
To explore the impact of the similarity threshold $\delta$, prior matching coefficient $\eta$ and optimal transport coefficient $\zeta$ on GCTM-OT, we conduct a parameter analysis on the \emph{`Bothering'} dataset using a 20-topic configuration.

For each parameter, we test five different values: $\delta \in \{0.2, 0.4, 0.6, $ $0.8, 1.0\}$, $\eta \in \{0.6, 0.8, 1.0, 1.2, 1.4\}$, and $\zeta \in \{0.6, 0.8, 1.0, 1.2, 1.4\}$, while keeping other parameters at their default settings. Detailed results are listed in Table~\ref{tab:Sensitivity_Analysis}. For clarity, we present only CAST’s results as the comparison baseline due to its competitive performance. We observe that our proposed GCTM-OT consistently outperforms the competitive CAST across various parameter combinations, demonstrating its robustness to these hyperparameters.

\subsection{Ablation Study}

To explore the impact of the transformer $\mathcal{T}$, large language models (LLMs) and different strategies (such as optimal transport, semantic-aware contrastive learning and goal-oriented text representation) on GCTM-OT, we conduct ablation studies on the \emph{`Bothering'} dataset using a 20-topic configuration.

\begin{table}[!t]
  \centering
    \resizebox{0.46\textwidth}{!}{
  \begin{tabular}{c|l|ccccc}
  \toprule
  \multicolumn{2}{c|}{\textbf{Model}} &  $\bm{C_P}$   &  $\bm{C_A}$   & $\bm{NPMI}$  &  $\bm{UCI}$  & $\bm{UT}$ \\
  \midrule
  \multicolumn{2}{c|}{CAST} & 0.2575  & 0.1592  & 0.0258  & -0.3314  & 0.6300  \\
  \midrule
  \multirow{5}[2]{*}{$\mathcal{T}$} & GCTM-OT (RoBERTa) & 0.3460  & 0.1944  & 0.0489  & 0.0414  & \textbf{0.9900} \\
        & {GCTM-OT(BERT)} & 0.2760  & 0.1673  & 0.0381  & 0.0639  & 0.9450  \\
         & {GCTM-OT(Sentence-BERT)} & 0.2760  & 0.1668  & 0.0371  & -0.0707  & 0.9500\\
        & GCTM-OT(MiniLM) & 0.3359  & 0.1939  & 0.0506  & 0.0312  & 0.9600  \\
        & GCTM-OT & \textbf{0.3518} & \textbf{0.2084} & \textbf{0.0524} & \textbf{0.0568} & 0.9800  \\
  \midrule
  \multirow{4}[2]{*}{Strategy} & {GCTM-OT w/o CL} &  0.2959  & 0.1957  & 0.0329  & -0.2770  & 0.9450  \\
        & GCTM-OT w/o GR & 0.3028  & 0.1993  & 0.0428  & -0.0584  & \textbf{0.9900} \\
        & GCTM-OT w/o OT & 0.3148  & 0.2017  & 0.0436  & -0.0586  & 0.9800  \\
        & GCTM-OT & \textbf{0.3518} & \textbf{0.2084} & \textbf{0.0524} & \textbf{0.0568} & 0.9800  \\
  \midrule
  \multirow{5}[2]{*}{LLM} & GCTM-OT(DeepSeek-V2.5)  & 0.3276  & 0.1928  & 0.0403  & -0.1403  & 0.9400  \\
        & GCTM-OT(GPT-4o-mini) & \textbf{0.3802} & \textbf{0.2182} & \textbf{0.0613} & \textbf{0.2174} & \textbf{0.9800} \\
        & GCTM-OT(Gemini) & 0.2704  & 0.1818  & 0.0413  & 0.1093  & 0.9050 \\
        & GCTM-OT(Mistral) & 0.3031  & 0.1836  & 0.0456  & 0.0824  & 0.9050   \\
        & GCTM-OT & 0.3518  & 0.2084  & 0.0524  & 0.0568  & 0.9800  \\
  \bottomrule
  \end{tabular}%
}
  \caption{Comparison results of ablation studies.}
\vspace{-0.4cm}
  \label{tab:Ablation_Study}%
\end{table}%

For the transformer $\mathcal{T}$ ablation, we utilize five variants: BERT~\cite{DBLP:conf/naacl/DevlinCLT19}, MiniLM~\cite{DBLP:conf/nips/WangW0B0020}, MPNet~\cite{DBLP:conf/nips/Song0QLL20},  Sentence-BERT~\cite{DBLP:conf/emnlp/ReimersG19} and RoBERTa~\cite{DBLP:journals/corr/abs-1907-11692}. For strategy ablation, we use variants of GCTM-OT that exclude optimal transport (GCTM-OT w/o OT), goal-oriented text representation (GCTM-OT w/o GR) and the semantic-aware mechanism in contrastive learning (GCTM-OT w/o CL). For LLM ablation, we explore various models: DeepSeek-V2.5~\footnote{\url{https://api-docs.deepseek.com/news/news0905}}, GPT-4o-mini~\footnote{\url{https://platform.openai.com/docs/models/gpt-4o-mini}}, GPT-3.5~\footnote{\url{https://platform.openai.com/docs/models/gpt-3.5-turbo}}, Gemini-1.5-pro~\footnote{\url{https://ai.google.dev/gemini-api/docs/models\#gemini-1.5-pro}} and Mistral~\footnote{\url{https://mistral.ai/news/mixtral-8x22b}}.

As shown in Table~\ref{tab:Ablation_Study}, GCTM-OT with different transformers, strategies and LLMs consistently outperforms the competitive baseline in terms of topic coherence and diversity metrics. For clarity, we present only CAST’s results as the baseline.


\section{Conclusion}

We introduce \emph{Human-TM}, a novel topic modeling task formulation that embeds human-provided goals into topic discovery. To tackle this task, we propose the GCTM-OT, a goal-prompted contrastive topic model with optimal transport to generate interpretable, diverse and goal-oriented topics. Experiments on three public subreddit datasets show that GCTM-OT consistently outperforms state-of-the-art baselines in coherence, diversity and goal alignment. By directly embedding user intent into the modeling process, \emph{Human-TM} reduces the need for manual filtering and improves the practical utility of topic models for targeted analysis. Future work will explore hierarchical extensions of \emph{Human-TM}, enabling the generation of topic hierarchies aligned with human-provided goals.

\section*{Acknowledgement}
This work is supported by the National Natural Science Foundation of China (Grant No. 62102192), the fellowship of China Postdoctoral Science Foundation (Grant No. 2022M710071), the Fundamental Research Funds for the Central Universities of China (Grant Nos. PA2025IISL0107 and aiia-24-01) and the Innovation and Entrepreneurship Program of Jiangsu Province (Grant No. JSSCBS20210530).


\bibliographystyle{ACM-Reference-Format}

\bibliography{reference}

\balance

\appendix









\end{document}